\documentclass[runningheads]{llncs}
\usepackage{graphicx}
\graphicspath{{figures/}}
\usepackage{comment}
\usepackage{amsmath,amssymb} 
\usepackage{color}
\usepackage[ruled,vlined]{algorithm2e}
\usepackage{multirow}

\usepackage{xspace}
\usepackage[pagebackref=true,breaklinks=true,letterpaper=true,colorlinks,bookmarks=false]{hyperref}
\usepackage[capitalise]{cleveref}

\makeatletter
\DeclareRobustCommand\onedot{\futurelet\@let@token\@onedot}
\def\@onedot{\ifx\@let@token.\else.\null\fi\xspace}

\def\eg{\emph{e.g}\onedot} 
\def\ie{\emph{i.e}\onedot} 
\def\cf{\emph{c.f}\onedot}

\def\etal{\emph{et al}\onedot}
\makeatother

\begin{document}
\pagestyle{headings}
\mainmatter
\def\ECCVSubNumber{3028}  

\title{Cluster-level Feature Alignment \\ for Person Re-identification} 

\titlerunning{Cluster-Level Feature Alignment}
%
\author{Qiuyu Chen\inst{1} \and
Wei Zhang\inst{2,3} \and
Jianping Fan\inst{1}}
\authorrunning{Q. Chen et al.}
%
\institute{ Department of Computer Science, UNC Charlotte \and
Shanghai Key Laboratory of Intelligent Information Processing \and School of Computer Science, Fudan University \\
\email{\{qchen12,jfan\}@uncc.edu}, \email{weizh@fudan.edu.cn}}
\maketitle

\begin{abstract}
Instance-level alignment is widely exploited for person re-identification, \eg spatial alignment, latent semantic alignment and triplet alignment.
This paper probes another feature alignment modality, namely cluster-level feature alignment across whole dataset, where the model can see not only the sampled images in local mini-batch but the global feature distribution of the whole dataset from distilled anchors. Towards this aim, we propose anchor loss and investigate many variants of cluster-level feature alignment, which consists of iterative aggregation and alignment from the overview of dataset. 
Our extensive experiments have demonstrated that our methods can provide consistent and significant performance improvement with small training efforts after the saturation of traditional training. In both theoretical and experimental aspects, our proposed methods can result in more stable and guided optimization towards better representation and generalization for well-aligned embedding.
\end{abstract}

\section{Introduction}
Person re-identification (ReID) is an essential component of intelligent computer vision systems. It has drawn increasing interest in many applications, such as surveillance, activity analysis and long-term tracking.  Given an image of a person-of-interest captured by one  camera, the goal is to re-identify this person from images captured by multiple cameras without overlapping viewpoints. As an instance-level recognition problem, the ReID task is inherently challenging. First, intra-class variations are typically huge due to significant changes of visual appearances caused by camera viewing conditions, human pose variations, occlusions, \etal. Second, the inter-class variations can be quite small because people may wear similar clothes.

\begin{figure}
    \centering
    \includegraphics[width=.44\linewidth]{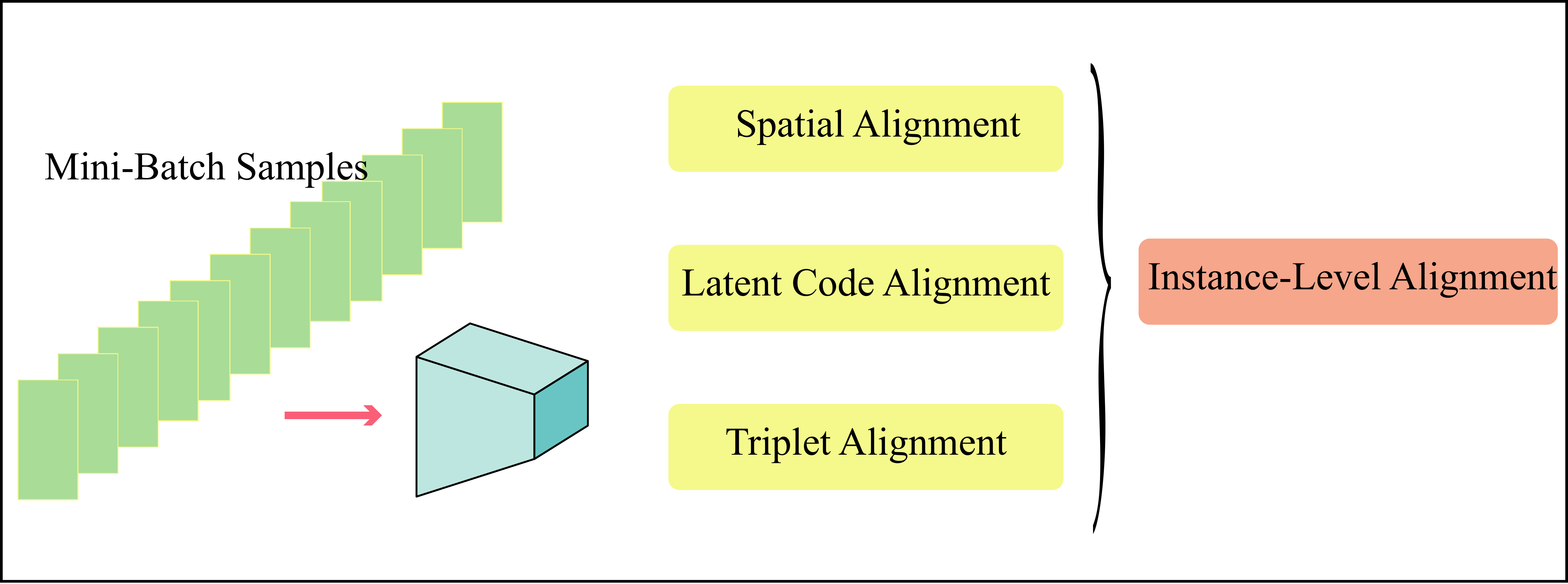}
    \includegraphics[width=.54\linewidth]{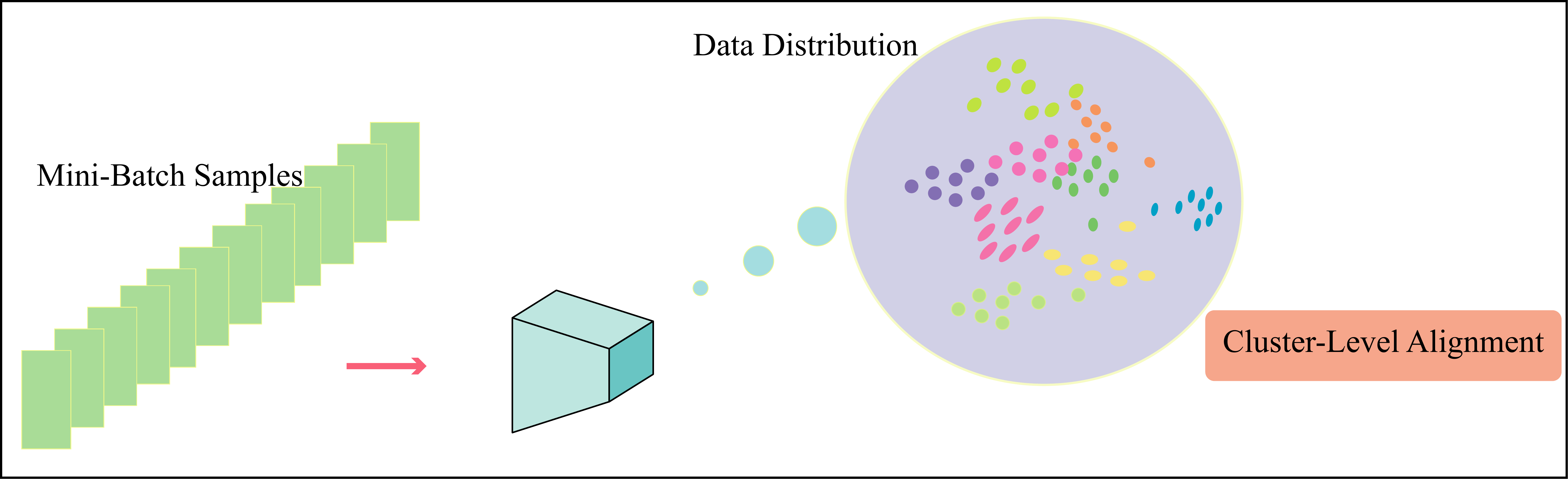}
    \vspace*{-0.38cm}
    \caption{From the view of alignment modality: instance-level alignment (left) and cluster-level alignment (right).}
    \label{fig:comp-alignment}
    \vspace*{-0.68cm}
\end{figure}
To address these challenges of  intra-class diversity and inter-class inseparability, a lot of efforts have been devoted to deep learning for its strong capability on discriminative feature extraction. Most of existing methods put the training process of person ReID under classification framework, where intermediate features are extracted to compute the similarity between query and gallery images during test. Some tailor-made neural networks are proposed to incorporate localization/attention or disentanglement for feature alignment. The former one aligns features in 2D spatial dimension and the latter one targets latent semantic alignment, but the essence of those approaches is instance-level alignment (\cf \cref{fig:comp-alignment} left).
In addition, a large variety of loss functions have been proposed for metric learning in person ReID. For example, the two most prevalent loss functions are classification loss, \eg cross entropy loss, and metric-learning based loss, \eg hard-negative mining triplet loss~\cite{hermans-triplethard}. For those advanced networks driven by such popular loss definitions, although successful, we argue that they can still be categorized as instance-level alignment (\cf \cref{fig:comp-alignment} left).
The intrinsic reason of such constraint is attributed to the adoption of general classification framework, where the interaction it builds can only dwell within the sampled mini-batch but cannot see more neighbors in the distribution of the whole dataset. As a result, it inhibits the growth of intra-class compactness and inter-class separability.

Aiming to break through the aforementioned limitations and step beyond the instance-level feature alignment, we propose a succinct and efficient method to enable cluster-level interaction in feature space, targeting the alignment from an overview of latent feature distribution (\cf \cref{fig:comp-alignment} right). ReID is in essential a metric learning problem. When projected to the learned feature space, feature points are expected to gather into compact clusters respecting their labels and such cluster-level interaction may inhabit better formulation of the clusters. We define the center of each feature cluster as the anchor. In a computational efficient manner, the anchors generated from aggregation serve as a supervision from the distribution of whole dataset and enable the model to see other training images in the dataset indirectly. In practice, after the saturation of traditional training, we manipulate two iterative steps to further intensify the cluster compactness: (a) Aggregate cluster features across dataset, stepping beyond the limitation of a classification framework; (b) Align features under the guidance from aggregated anchors. We claim such cluster-level feature alignment is much more promising for identity-related representation.

\begin{figure}
    \vspace*{-0.38cm}
    \centering
    \includegraphics[width=\linewidth]{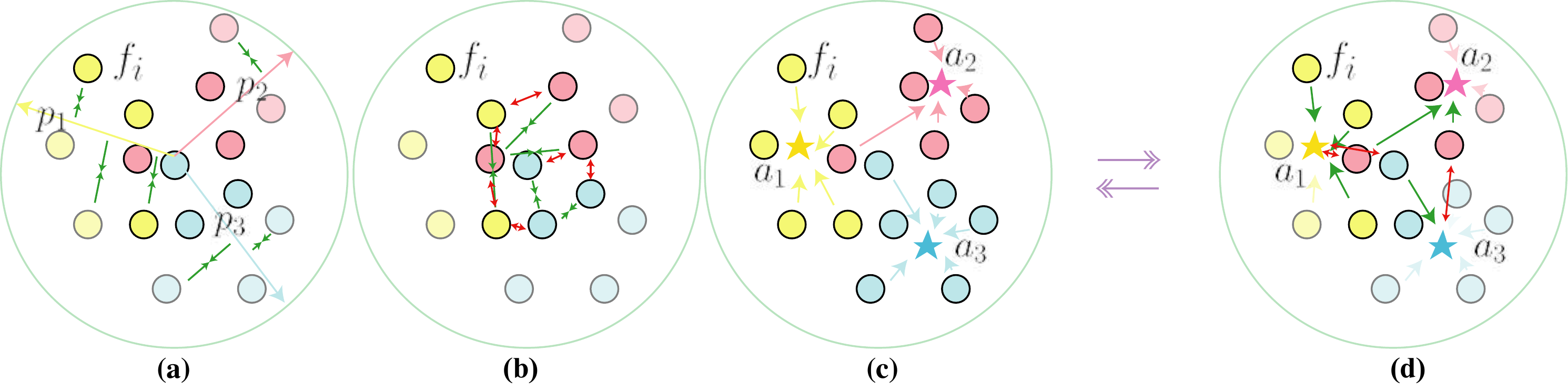}
    \vspace*{-0.68cm}
    \caption{From the view of optimization: Anchor loss provides more stability and consistency for optimization. (a) Classification loss pulls feature sample $f_i$ towards the corresponding classifier vector $p_j$ in fully connected layer; (b) Triplet loss probes the interaction within sampled mini-batch (denoted as solid color); (c-d) Anchor loss enables the sampled mini-batch to see the anchors aggregated from all the siblings through iterative aggregation(c) and alignment(d).}
    \label{fig:view-optimization}
    \vspace*{-0.68cm}
\end{figure}

Besides the view of feature alignment modality (\cf \cref{fig:comp-alignment}), the proposed method, called anchor loss, provides consistent optimization for metric learning which benefits training as well as generalization process. 
The classification loss tries to align the features in orders according to the classification labels. Specifically, the inner product $\textbf{f}^{\top}_{i}\textbf{p}_j$ between classifier $p_{j}$ in the full-connected layer and the feature vector $f_i$ is increased as $p_{j}$ and $f_i$ pulling towards each other (\cf \cref{fig:view-optimization}(a)). It has promising convergence but unnecessary penalize the intra-class variance if classifier diversifies channel-wise focus on decision. Also, it handles training mini-batch samples by simply averaging individual losses, thus can only build sample connection from the identity implicitly. 
Triplet loss~\cite{hermans-triplethard} tries to align features in more explicitly way. It optimizes the intra-class and inter-class distance by mini-batch interaction with proper sampling. From \cref{fig:view-optimization}(b), we can see that the optimization direction of triplet loss is highly dependent on the mini-batch sampling and inevitably introduces uncertainty and inconsistency. On the other hand, anchor loss enables the sampled mini-batch to see the anchors $\textbf{a}_{j}$ aggregated from all the siblings, bearing more consistency (\cf \cref{fig:view-optimization}(c-d)). Anchors generated from aggregation provide strong guidance and propagate the global information from the distribution of dataset to local mini-batch training.  
Based on extensive experiments, we demonstrate that anchor loss can consistently boost the generalization.

Overall, in this proposed paper, we learn a metric to overcome intra-class diversity and inter-class confusion for person ReID based on anchor-based min-batch training. Although each mini-batch of samples is a small subset of the dataset, we can successfully capture the global information during training through anchors which play key roles in the proposed cluster-level feature alignment. A small number of representative anchors propagate rich knowledge from the distribution of dataset into local training batch in a computational efficient manner.

\section{Related Work}\label{sec:related_work}

\textbf{Person Re-Identification}
A large group of person re-identification network focuses on feature alignment.
In general, there are two kinds of feature alignment: (a) spatial feature alignment by attention and localization (b) latent feature alignment by disentanglement.

In spatial feature alignment, it can be categorized as self-supervised and extra-supervised methods. We consider hand-crafted splitting as one representation of self-supervision. Sun \etal~\cite{sun-pcb} propose PCB to split the intermediate features horizontally in order to align the feature in local spatial parts and is widely used by \cite{fu-hpyramid,guo-part,quan-autopart,sun-perceive,zheng-pyramidal}.
Quite some works \cite{jiehu-squeeze,dangweili-cvpr2017,shuangli-cvpr2018,weili-cvpr2018,jingxu-cvpr2018} proposed similar and effective part-aligning CNN networks for locating context regions and then extract these regional features for ReID.
Extra-supervision leverages human part detector \cite{haiyu-cvpr2017}, human pose \cite{liu-pose} or human body parsing \cite{kalayeh-semanticpart} to provide more accurate localization. For example, \cite{miao-pose,saquib-pose,subramaniam-pose,xu-pose} incorporate external pose attention maps to align the feature in deformable spatial space of human body. SPReID \cite{kalayeh-semanticpart} utilizes a parsing model to generate five different predefined human part masks to compute more reliable part representations, which achieves promising results on various person ReID benchmarks. Dense semantic alignment \cite{zhang-densely} went one step further, it addressed the body misalignment by leveraging the estimation of the dense semantics of a person image, and constructed a set of densely semantically aligned part images for re-identification. Other methods for spatial alignment include the attention from attribute~\cite{tay-aanet}, forground mask~\cite{tian-bgmask}, \etal.

For latent feature alignment, DG-net~\cite{zheng-joint} proposed a disentanglement solution by GAN and Autoencoder to decouple input into appearance code and structure code, and extract pose-invariant features. Whereas, there has been a growing interest in using generative models to augment training data and enhance the invariance to input changes \cite{zhedong-iccv2017,xiang-eccv2018,yan-tip2018}.

\noindent \textbf{Metric Learning: Center Loss and Triplet Loss}
In end-to-end learning process, several methods propose to explore iteration within mini-batch for feature alignment .
Zheng \etal~\cite{zheng-verificationloss} propose a verification loss to align the pairwise features.
Hermans \etal~\cite{hermans-triplethard} target the triplet samples and point out the triplet loss on hard examples mining is superior to batch-all triplet loss.
In face recognition, parametric center loss~\cite{wen-centerloss} is proposed to align the intra-class distance and it is used by \cite{luo-bnneck} for person reID, which looks similar to our proposed anchors.
Whereas, our motivation is essentially different.
Center loss treats the parametric center as an auxiliary decision factor similar to the classifier $p_{j}$ in \cref{fig:view-optimization}(a), which is jointly optimized under classification framework and only builds the connection from identity label implicitly.
On the contrary, our proposed anchor loss complies the embedded feature distribution and distills the knowledge from sibling samples (\cf \cref{fig:view-optimization}(c)) to enable the interaction in cluster level explicitly.
More recently, Wen \etal~\cite{wen2019comprehensive} revisit center loss and propose to use the classifier layer as the center for each class, which further validates our motivation difference.
Moreover, it suffers from large instability due to random initialization for parametric center.
On the other hand, our method provides constant improvement because the aggregation distills knowledge from dataset distribution.

In summary, those methods never push towards the constraint of classification framework in mini-batch training and scrutinize the modality in cluster-level feature alignment.
Probably due to the concern about training efficiency, cross-dataset aggregation is not fully investigated in deep CNN methods.
However, we conduct a comprehensive study on different variants of the cluster-level interaction, which has demonstrated our method could be trained effectively and efficiently with small training efforts.

\section{Proposed Method}

Person ReID aims to establish the identity correspondences between each query image and gallery images across different cameras. We use Convolutional neural network (CNN) to extract image features due to its strong representation power.  To learn discriminative representation that is  robust against intra-class variation and interclass confusion, we take advantage of three different loss functions to train the model: (1) cross-entropy classification loss $L_{cls}$ (\cref{fig:view-optimization}(a)); (2) triplet loss $L_{trip}$ (\cref{fig:view-optimization}(b)); (3) anchor loss $L_{anchor}$ (\cref{fig:view-optimization}(d)).
$L_{cls}$ and $L_{trip}$~\cite{hermans-triplethard} are widely used for person ReID, and they are both instance-level optimization as illustrated in \cref{fig:view-optimization}(a-b).  To  propagate rich knowledge from outside samples into each mini-batch, the designed anchor loss $L_{anchor}$ targets cluster-level supervision which has two options:

(a)  Anchor Loss for Intra-Class Compactness: $L_{anchor}$ pulls the feature vector towards the anchor where its label belongs:
\vspace*{-0.28cm}
\begin{equation}\label{eq:intra-anchor}
L_{anchor} = \frac{1}{|\mathcal{B}|} \sum_{i\in \mathcal{B}} \sum_{j=1}^{C} \delta(y_{i}=j) D(\textbf{f}_{i}, \textbf{a}_{j})
\end{equation}
where $|\mathcal{B}|$ denotes the number of samples in the mini-batch $\mathcal{B}$, and $C$ is the number of classes.  $\textbf{f}_{i}$ and $y_{i}$ are the feature vector and the label of the sample $i$ in $\mathcal{B}$, respectively.
 $\textbf{a}_{j}$ is the anchor for the $j-$th class. 
 $D(\textbf{f}_{i}, \textbf{a}_{j})$ is the distance between the sample $\textbf{f}_{i}$ and the anchor $\textbf{a}_{j}$.
 $\delta(y_{i}=j)=1$ if the condition is satisfied, \ie the label of sample $i$ in the mini-batch equals $j$; otherwise, $\delta(y_{i}=j)=0$.
 
(b) Triplet Anchor Loss for Intra-Class Compactness $\&$ Inter-Class Separability: Motivated by hard sample mining~\cite{hermans-triplethard}, we add extra inter-class penalty to target the hard/confused anchor mining:
\vspace*{-0.28cm}
\begin{equation}\label{eq:triplet-anchor}
L_{TripletAnchor} = \frac{1}{|\mathcal{B}|} \sum_{i\in \mathcal{B}}\sum_{j=1}^{C} \delta(y_{i}=j) [D(\textbf{f}_{i}, \textbf{a}_{j}) - \underset{k \neq j}{min} D(\textbf{f}_{i}, \textbf{a}_k) + margin]
\end{equation}
Triplet anchor loss not only pulls the samples to the anchor in the same class close, but also pushes the negative samples further away than the distance between anchor and positive samples. It could be more discriminative by simultaneously taking into account intra-class compactness and inter-class separability. On the other hand, it may import inconsistency during the optimization in a similar way as triplet loss (\cf \cref{fig:view-optimization}(b)).

\subsection{Two-Staged Training}

\begin{figure}
    \vspace*{-0.38cm}
    \centering
    \includegraphics[width=\linewidth]{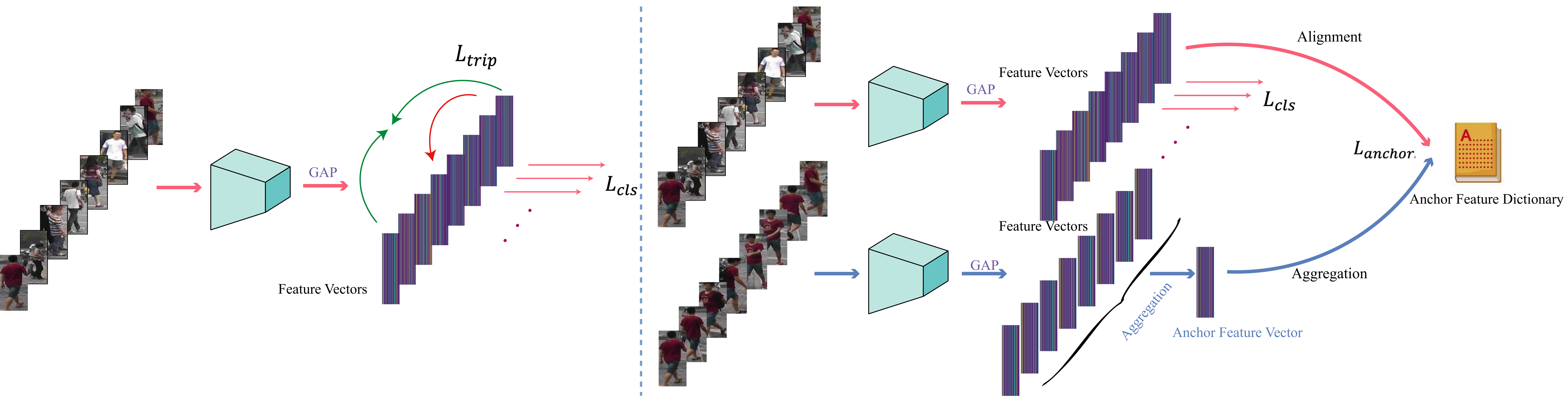}
    \vspace*{-0.58cm}
    \caption{Two training stages of the proposed metric learning framework. Stage I (left): instance-level alignment with $L_{cls} + L_{trip}$. Stage II (right): feature aggregation respecting class label to generate anchors, and cluster-level alignment with $L_{cls} + L_{anchor}$.}
    \label{fig:overview}
    \vspace*{-0.58cm}
\end{figure}

As shown in \cref{fig:view-optimization}, anchors generated from aggregation contribute more consistent optimization and less noisy guidance during training process.
However, it is based on the assumption that distribution of embedded features is approximately cluster-formed.
During the early training phase when the feature distribution is still random and stochastic, such aggregated anchors may contain misleading information and impair the training process.
  The cluster-level supervision could be more effective after the saturation of traditional training stage, and therefore our training consists of two stages: 

\textbf{Stage I:} Train the model in the traditional manner with loss function $L=L_{cls} + L_{trip}$, cultivating the initial formulation of clusters (\cf \cref{fig:overview}(left));

\textbf{Stage II:} Train the model under the cluster level supervision with loss function $L=L_{cls} + L_{anchor}$, capturing distribution of whole dataset in embedded feature space (\cf \cref{fig:overview}(right)).

\vspace*{-0.18cm}

\subsection{Generation and Update of Anchors}
During the learning process, we take the anchors as the \textbf{global} supervision from data distribution to align feature towards a better representative embedding in \textbf{local} mini-batch training, \ie pushing towards the target anchor (\cref{eq:intra-anchor}) and pulling samples away from the confusion anchor (\cref{eq:triplet-anchor}). Aggregation and alignment are iteratively performed  to further reduce the intra-class variance and inter-class entanglement.  

\vspace*{-0.48cm}
\subsubsection{Generation}
When the latent features are extracted over training dataset respecting their labels, we consider two approaches to estimate optimal anchors during aggregation:

(a) Average aggregation: When the embedded features are approximately cluster-formed, aggregate the embedded features $f_{i}$ for each class in training dataset $\mathcal{T}$:
\begin{equation}\label{eq:avg_aggregation}
\textbf{a}_j =  \frac{\sum_{i\in \mathcal{T}} \delta(y_{i}=j)\textbf{f}_{i} }{\sum_{i\in \mathcal{T}}\delta(y_{i}=j)}
\end{equation}

(b) Voting by confidence: Taking the prediction probability $P(j|i)$ for class $j$ as the contribution, aggregate the embedded features by a weighted mean:
\begin{equation}\label{eq:weighted_aggregation}
\textbf{a}_j=\frac{\sum_{i\in \mathcal{T}} \delta(y_{i}=j)  P(j|i)  \textbf{f}_{i}}{\sum_{i\in \mathcal{T}}\delta(y_{i}=j) P(j|i)}
\end{equation}

\cref{eq:avg_aggregation} treats each sample's contribution to anchor equally and could be an effective estimation to eliminate the variance of latent feature distribution caused by pose, camera view condition, background, \etal. It may work well supposing the training samples are equally distributed in terms of variance.
\cref{eq:weighted_aggregation} takes the classifier confidence as the contribution and help to revealing the early portrait of anchors. Intuitively, when the feature cluster distribution is still stochastic in the early training stage, easy samples, which may contains less noise and variance thus converges faster, could be close to optimal anchors and guide the hard samples moving towards the estimated optimal centers.

\scalebox{.62}{
\begin{minipage}{.5\linewidth}
    \begin{algorithm}[H]
        {
            \caption{Fix Anchors}
            \label{alg:anchors-fix}
            \For{$i \gets 1$ to \textbf{StartEpoch} $E_{start}$ } {
                Update model parameter $W$ by $L=L_{trip}+L_{cls}$;
            }
           {\color{blue} Update $\textbf{c}_j$ for each class $j$\;}
            \For{$i \gets E_{start}$ to \textbf{EndEpoch} $E_{end}$ } {
                \For{each iteration}{
                    Update model parameter $W$ by $L=L_{anchor}+L_{cls}$\;
                }
            }
        }
    \end{algorithm}
\end{minipage}
}
\scalebox{.6}{
\begin{minipage}{.5\linewidth}
    \begin{algorithm}[H]
        {
            \caption{Update Anchors Each Epoch}
            \label{alg:anchors-epoch}
            \For{$i \gets 1$ to \textbf{StartEpoch} $E_{start}$ } {
                Update model parameter $W$ by $L=L_{trip}+L_{cls}$;
            }
            \For{$i \gets E_{start}$ to \textbf{EndEpoch} $E_{end}$ } {
                \For{each iteration}{
                    Update model parameter $W$ by $L=L_{anchor}+L_{cls}$\;
                }
                {\color{blue} Update $\textbf{c}_j$ for each class $j$};
            }
        }
    \end{algorithm}
\end{minipage}
}
\scalebox{.6}{
\begin{minipage}{.5\linewidth}
    \begin{algorithm}[H]
        {
            \caption{Update Anchors Each Iteration}
            \label{alg:anchors-iteration}
            \For{$i \gets 1$ to \textbf{StartEpoch} $E_{start}$ } {
                Update model parameter $W$ by $L=L_{trip}+L_{cls}$;
            }
            \For{$i \gets E_{start}$ to \textbf{EndEpoch} $E_{end}$ } {
                \For{each iteration}{
                    Update model parameter $W$ by $L=L_{anchor}+L_{cls}$\;
                    {\color{blue} Update $\textbf{c}_j$ for each class $j$};
                }
            }
        }
    \end{algorithm}
\end{minipage}
}
\vspace*{-0.68cm}
\subsubsection{Update Frequency}

Ideally, we should calculate anchors by either \cref{eq:avg_aggregation} or \cref{eq:weighted_aggregation} after each forward and backward process when the training parameters are updated.
However, such a process is unrealistic in terms of training efficiency and thus we consider three options for the update frequency:

 \textbf{(a)} Constant (\cref{alg:anchors-fix}): When the model is trained until initial convergence, the cluster-level feature aggregation is calculated and serves as the fixed anchors during the following fine-tuning process;

 \textbf{(b)} Each Epoch (\cref{alg:anchors-epoch}): When the model is trained until some epoch, $E_{start}$, the anchors $\textbf{c}_j$ are updated after each following training epoch by either \cref{eq:avg_aggregation} or \cref{eq:weighted_aggregation};

 \textbf{(c)} Each Iteration (\cref{alg:anchors-iteration}): In each iteration trained with anchor loss, the anchor for class $j$ is updated as:
\begin{equation}
\textbf{a}_j^{t+1} = [1-\eta \cdot \sum_{i\in \mathcal{B}} \delta(y_{i}=j)] \cdot \textbf{a}_j^{t} + \eta \cdot \sum_{i\in \mathcal{B}} \delta(y_{i}=j) \textbf{f}_{i}
\end{equation}
where $\textbf{a}_j^{t}$ and $\textbf{a}_j^{t+1}$ are  the anchors for class $j$ at the $t-$th and $t+1-$th iterations respectively.  
To approximate the anchors calculated per iteration, we design the weight as $\eta = \frac{1}{\sum_{i\in \mathcal{T}}  \delta(y_{i}=j)}$ where $\mathcal{T}$ is the total training dataset.

Option (a) take the anchors calculated from the initial convergence as the optimal one and only one aggregation process is performed. It is based on the observation that the clusters are almost formed after the convergence of \textbf{Stage I} training and thus may provide stable optimization.
Option (b) adaptively updates the anchors after each training epoch in a manner similar to EM optimization: Estimate the anchor location according to the current feature cluster distribution in aggregation step and maximize the cluster compactness in alignment step.
Option (c) is a trade-off option between training efficiency and adaptive estimation for anchors, which can be viewed as an approximate approach to option (b).
From our experiments demonstrated later, we show this method could have comparable performance with option (b) and thus provides an alternative when the size of training samples are large or in the context of online learning.

\section{Experiments and Analysis}

\vspace*{-0.18cm}
\subsubsection{Experiment Setup}
We adopt the bag of tricks proposed by \cite{luo-bnneck}, \ie warm-up learning rate scheduler, random erasing augmentation~\cite{zhong2017random}, label smoothing, no stride down-sampling in last bottleneck of ResNet50 and bnneck (one additional batch normalization layer after classifier).
We use $L_2$ distance for the anchor loss and its variants, which benefits stable training.
We experiment our methods on three datasets, Market1501~\cite{zheng-market}, DukeMTMC-ReID~\cite{ristani-dukemtmc} and CUHK03~\cite{li-cuhk}. Market-1501 have 12,936 training images with 751 different identities. Gallery and query sets have 19,732 and 3,368 images respectively with another 750 identities. DukeMTMC-ReID includes 16,522 training images of 702 identities, 2,228 query and 17,661 gallery images of another 702 identities. CUHK03-NP is a new training-testing split protocol for CUHK03, it contains two subsets which provide labeled and detected (from a person detector) person images. The detected CUHK03 set includes 7,365 training images, 1,400 query images and 5,332 gallery images. The labeled set contains 7,368 training, 1,400 query and 5,328 gallery images respectively. The new protocol splits the training and testing sets into 767 and 700 identities. We note that our method is only used during the training stage and the evaluation methods stay the same with previous approaches.

Firstly, we present the experimental comparison without triplet loss to analyze three factors: starting time, aggregation methods and anchor loss functions. 
Secondly, we delve into the effects of aggregation anchors from a reconstruction experiments.
Thirdly, the comparison, when triplet loss is incorporated in the \textbf{Stage I} training until convergence, will be further analysed.
Lastly, we demonstrate the advantages of our method comparing to the parametric center loss~\cite{wen-centerloss}.

\vspace*{-0.18cm}
\subsection{Ablation Study for Three Factors}\label{sec:ablation}

\begin{figure}
    \centering
    \includegraphics[width=\linewidth]{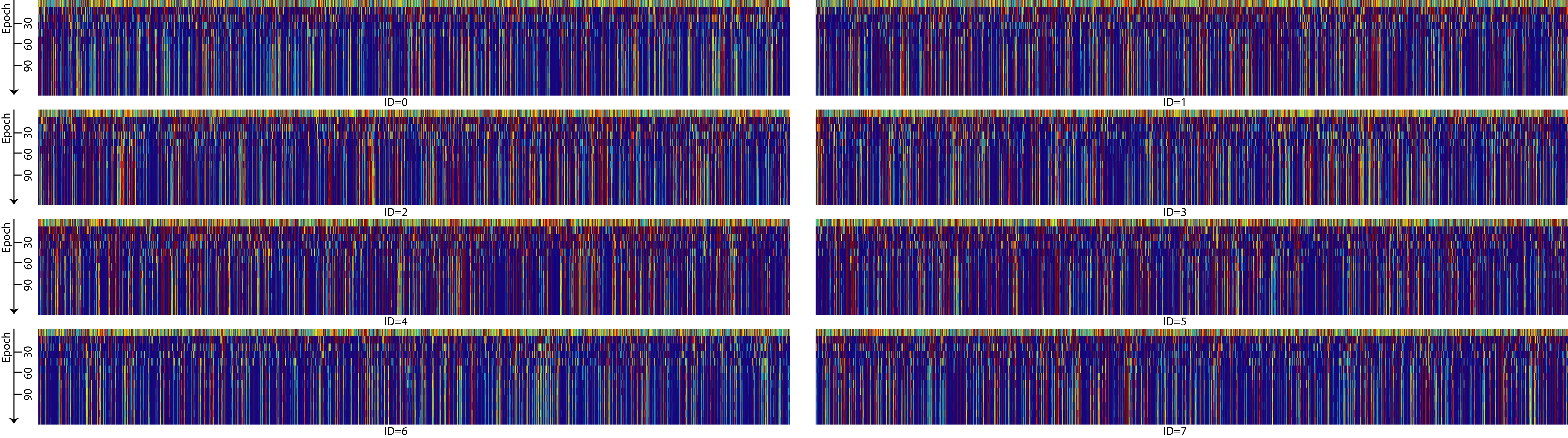}
    \caption{The change of sampled anchors (id = {0,1,...,7}) along training epochs (\ie 0, 10, 20, ... 120): Each image corresponds to an ID ({0,1,...,7}). Each row in image represents the anchor feature vector (2048 dimension) in the sampled epoch. Total 13 epochs from 0 to 120 with step of 10 is sampled. Zoom in to see the details. The sampled anchors are calculated from the checkpoints of training without anchor loss.}
    \label{fig:id-feature-change}
    \vspace*{-0.68cm}
\end{figure}

From the experimental results in \cref{tab:comp-ablation}, we conclude the impacts of three factors:

\noindent \textbf{(a) When to start} Attributed to better cluster distribution, stating aggregation and alignment after the convergence of initial training results in better generalization.
From \cref{fig:id-feature-change}, one can see the anchors change rapidly during the early training phase.
The transition becomes steady as the training process towards saturation.
It is in line with our analysis that anchor loss may impose unexpected prior and abet densely distributed clusters when applied early during training, impairing the generalization consequently.

\begin{table}[t]
    \centering
    \resizebox{\linewidth}{!}{
        \begin{tabular}{c|c c c|c c c}
            \hline
            \multirow{2}{*}{$E_{start}$} & \multicolumn{3}{c}{Rank@1} & \multicolumn{3}{c}{mAP} \\
            & $L_{Anchor}(f, y, a_{avg})$ & $L_{Anchor}(f, y, a_{weighted})$ & $L_{TripletAnchor}(f, y, a_{avg})$
            & $L_{Anchor}(f, y, a_{avg})$ & $L_{Anchor}(f, y, a_{weighted})$ & $L_{TripletAnchor}(f, y, a_{avg})$ \\ \hline
            - &  93.79\% &  93.79\% & 93.79\% & 84.69\% & 84.69\% & 84.69\% \\ \hline
            0 & 93.85\% & \textbf{94.06\%} & 83.86\% & 83.93\% & \textbf{84.95\%} & 67.48\% \\
            10 & \textbf{93.32\%} & 93.29\% & 93.29\% & \textbf{83.77\%} & 83.59\% & 83.65\% \\
            40 & 93.97\% & 93.91\% & \textbf{94.09\%} & \textbf{85.49\%} & 85.45\% & 85.43\% \\
            70 & 94.09\% & 94.09\% & \textbf{94.15\%} & 85.75\% & \textbf{85.89\%} & 85.81\% \\
            120 & \textbf{\color{blue}94.18\%} & 94.03\% & 94.09\% & \textbf{\color{blue}85.98\%} & 85.96\% & 85.90\% \\
            \end{tabular}
        }
    \caption{Ablation study for three factors on Market1501 dataset: starting epoch $E_{start}$, aggregation methods (\cref{eq:avg_aggregation}\&\cref{eq:weighted_aggregation}) and anchor loss choices(\cref{eq:intra-anchor}\&\cref{eq:triplet-anchor}). $f$ and $y$ denote the extracted feature and its label. Before $E_{start}$ $L=L_{cls}$ is used in the first stage training. Afterwards, either $L=L_{cls}+L_{Anchor}$ or $L=L_{cls}+L_{TripletAnchor}$ is applied. We updates the anchors each epoch as in \cref{alg:anchors-epoch}.}
    \label{tab:comp-ablation}
    \vspace*{-0.68cm}
\end{table}

\begin{figure}[!b]
    \centering
    \includegraphics[width=0.8\linewidth]{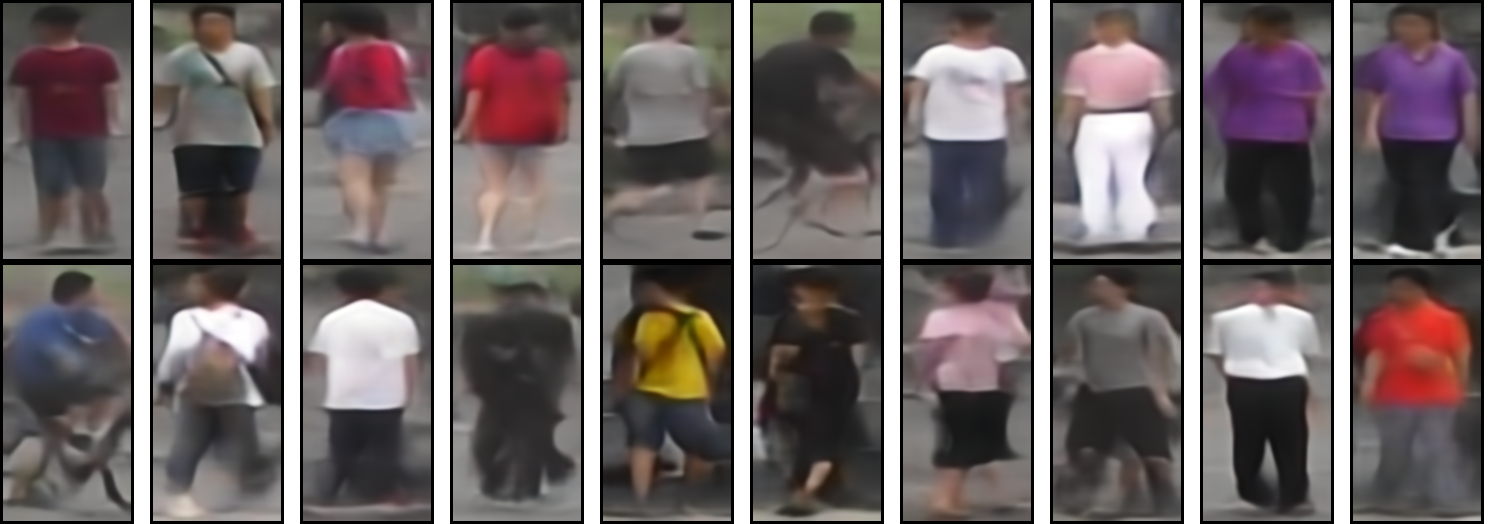}
    \vspace*{-0.38cm}
    \caption{Reconstruction results of anchors in the training dataset (ID={0,1,2,3...,19}).}
    \label{fig:recon-anchor}
\end{figure}

\noindent\textbf{(b) How to calculate anchors} When the feature alignment is still stochastic, \ie early training phase, calculating anchors with probability contribution (\cref{eq:weighted_aggregation}) produces better results.
Easy samples, which converge earlier, may contain less noises and variance, revealing the approximation of optimal anchors, \eg the anchors generated from reconstruction in \cref{fig:recon-anchor}.
After the initial training approaches convergence, the benefits of voting by confidence become less significant.

\begin{figure}[t]
    \vspace*{-0.38cm}
    \centering
    \includegraphics[width=0.8\linewidth]{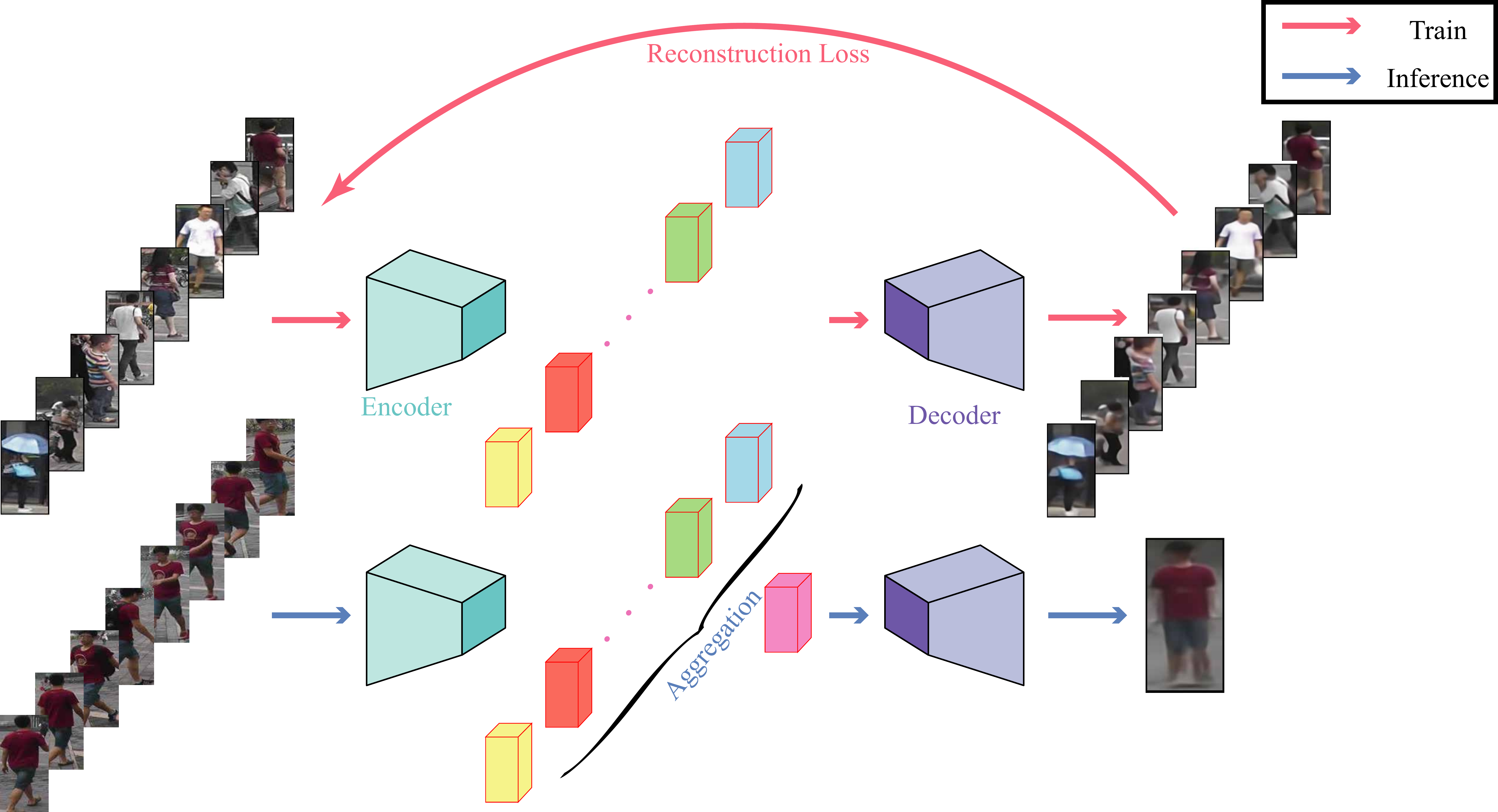}
    \caption{Reconstruction pipeline of anchors in image space: Encoder is transfered and frozen during the training of decoder. During the inference, the decoder reconstructs the anchor into image space.}
    \label{fig:recon-pipeline}
\end{figure}

\noindent\textbf{(c) How to look at anchors} Triplet anchor loss and intra-class anchor loss achieve comparable results when starting the aggregation and alignment in the intermediate stage of training convergence.
After the initial training saturation, intra-class anchor loss performs slightly better, implying the consistency weights more in training \textbf{Stage II}.
In experiments, we also find triplet anchor loss needs a proper tuning for the margin hyper-parameter.
Training is unstable when set to a high margin while easily saturated when set to a low one.
The best results are found when $margin=0$, where however training converges slower comparing to intra-class anchor loss.
It validates our initial assumption that, when the anchors are well aggregated after \textbf{Stage I} training convergence, optimization with consistency provides more benefits for generalization.

\vspace*{-0.18cm}
\subsection{What Would the Anchors Look Like}

In order to further validate our assumption about the benefits of aggregation, we train a decoder network to reconstruct the images from the feature maps before GAP (global average pooling), where encoder is the well-trained model without anchor loss.
A decoder structure slightly modified from DG-Net~\cite{zheng-joint} is used.
Then we generate anchors in image space by feeding the average aggregated feature maps before GAP into trained decoder (\cf \cref{fig:recon-pipeline}).
We note that GAP applied anchor feature map produce the same as anchor feature vector and 2D maps are used to preserve the spatial information for better reconstruction.
This anchor feature vector can be taken as the initial anchor during the start of training \textbf{Stage II}.
As the results in \cref{fig:recon-anchor}, the anchors could be a feature vector which dissects the view-variance, pose-variance as well as background-variance.
Comparing to the sample images from same class (\cf \cref{fig:recon-pipeline}), anchors generated from aggregation acts like a implicit regularization to remove noise and variance.
It complies that the average aggregation distills the constitutional id-related features over the sampling distribution.

\vspace*{-0.18cm}
\subsection{Further Discussion}

\begin{figure}[t]
    \vspace*{-0.38cm}
    \centering
    \includegraphics[width=.30\linewidth]{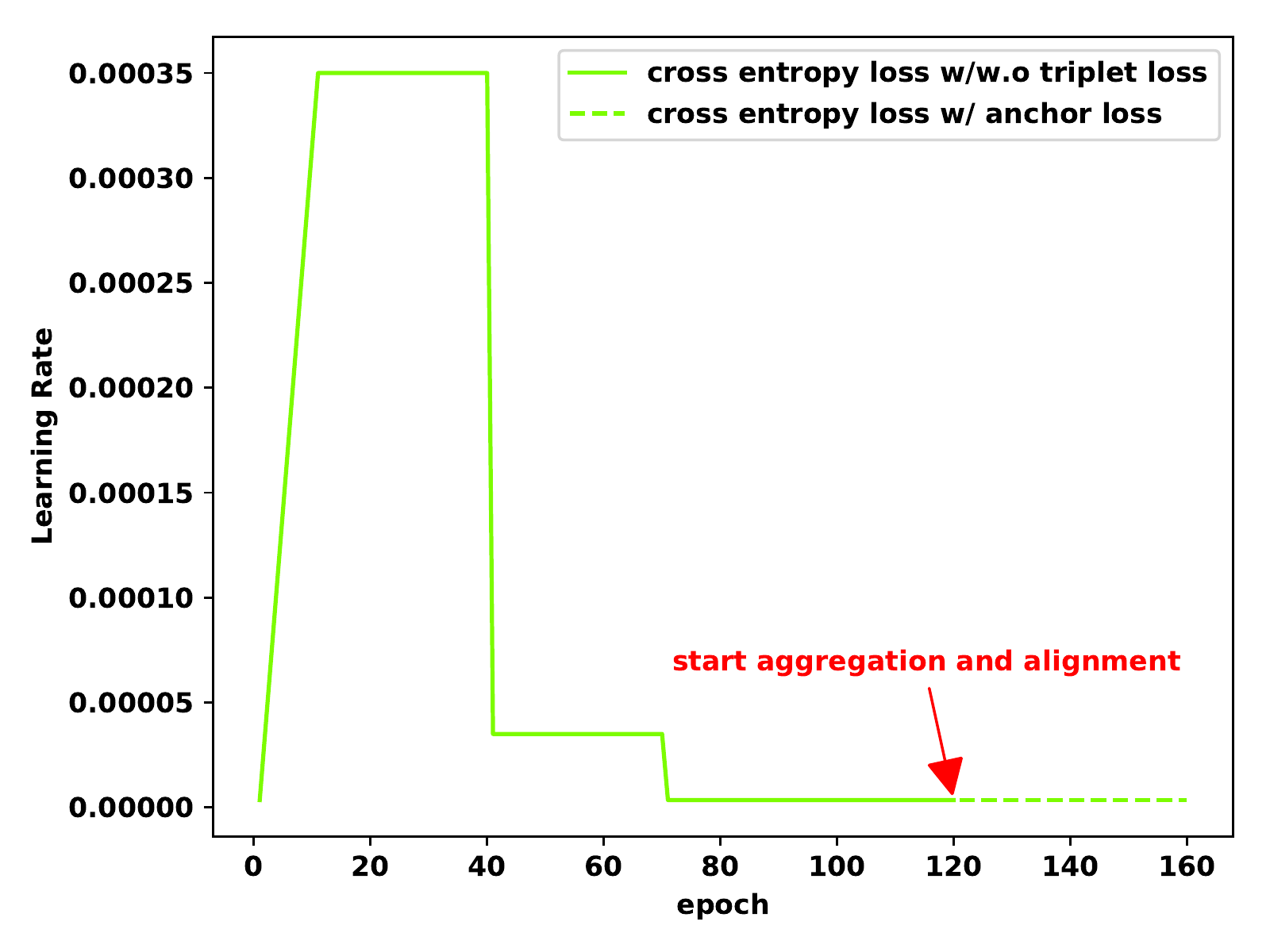}
    \includegraphics[width=.30\linewidth]{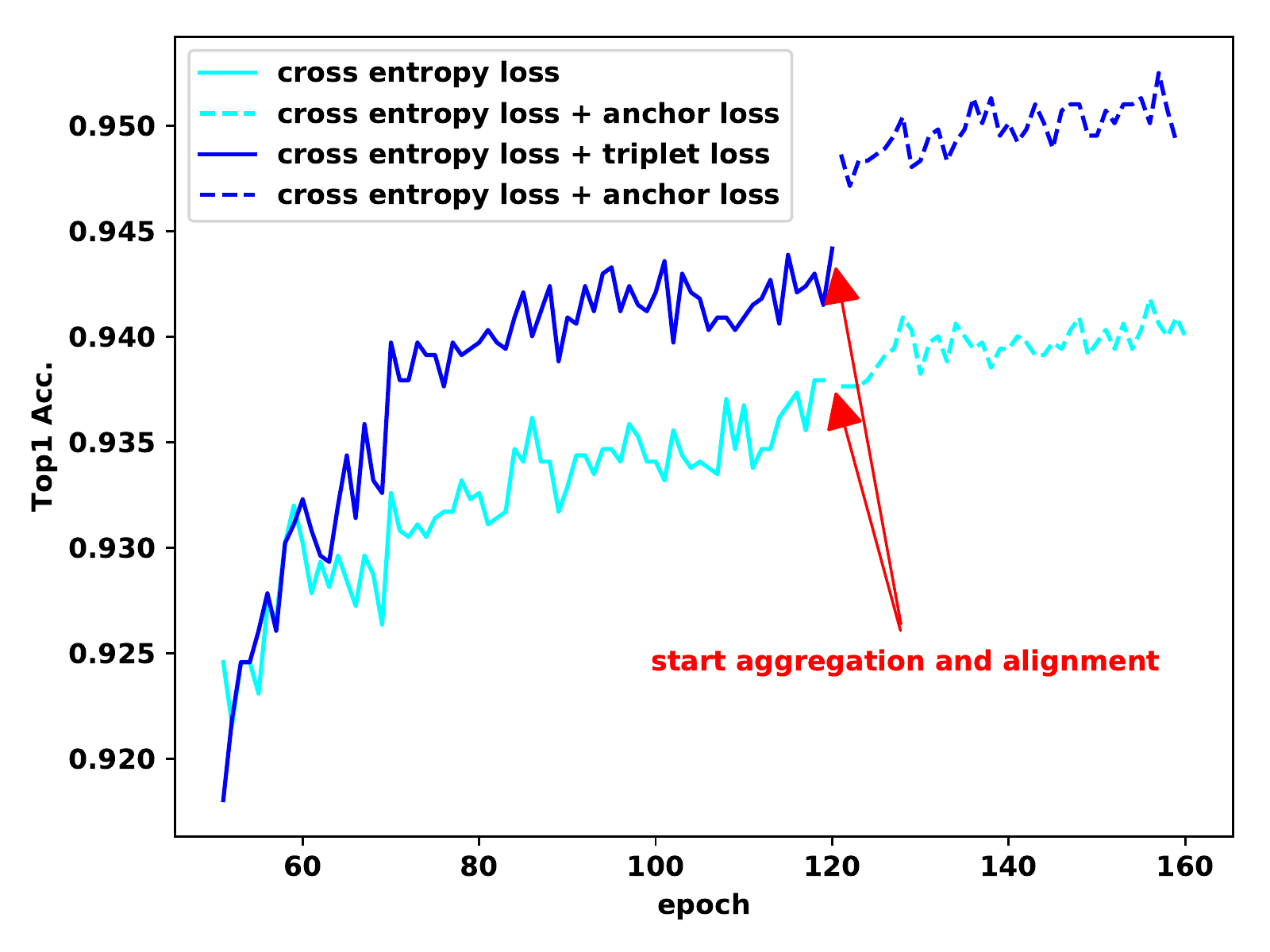}
    \includegraphics[width=.30\linewidth]{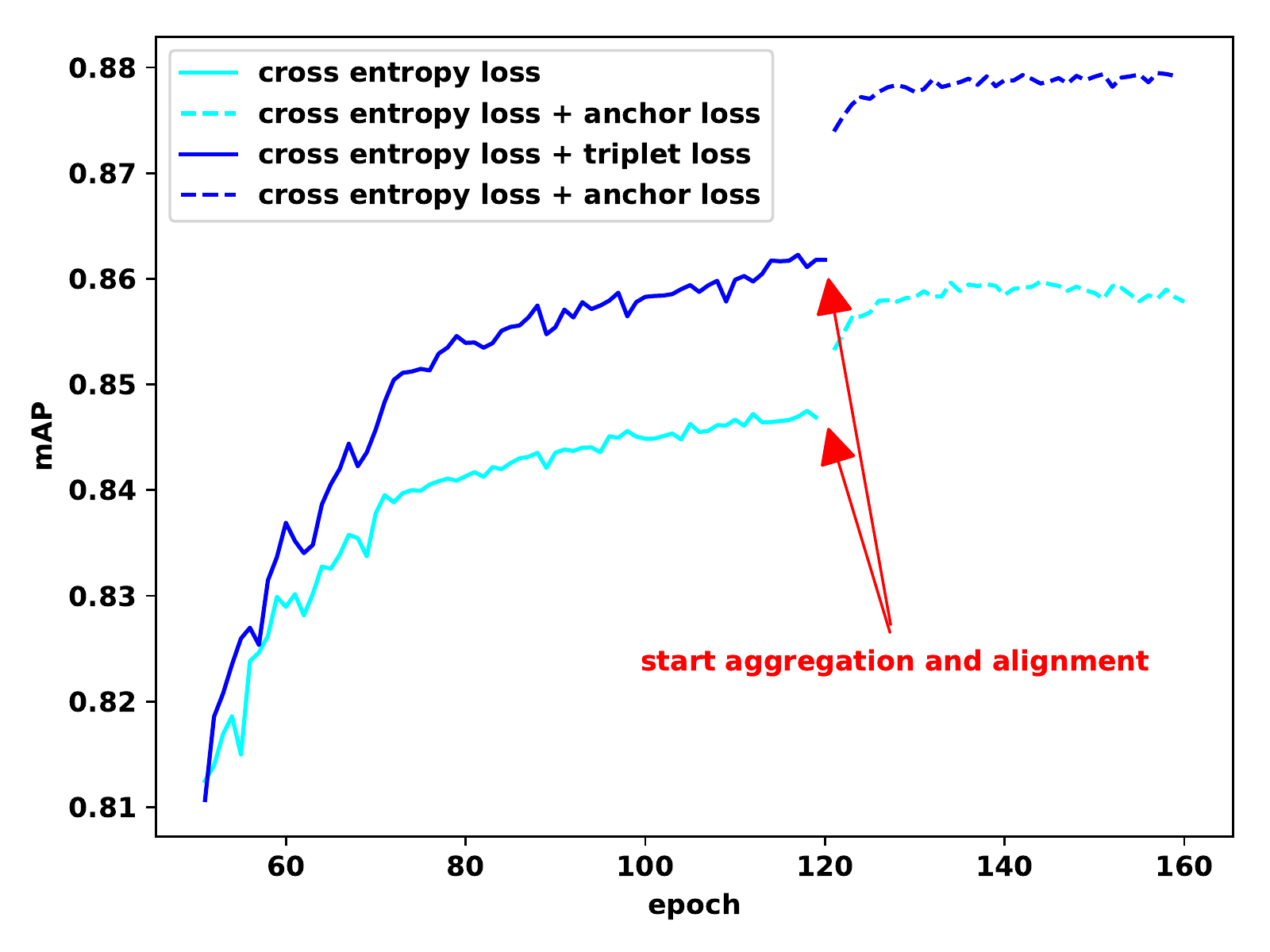}
    \vspace*{-0.38cm}
    \caption{Test result of anchor loss from checkpoints: Without bells and whistles, anchor losses boost the performance significantly after the initial training is saturated.}
    \label{fig:learning-curve}
    \vspace*{-0.58cm}
\end{figure}

\begin{figure}[t]
    \centering
    \includegraphics[width=\linewidth]{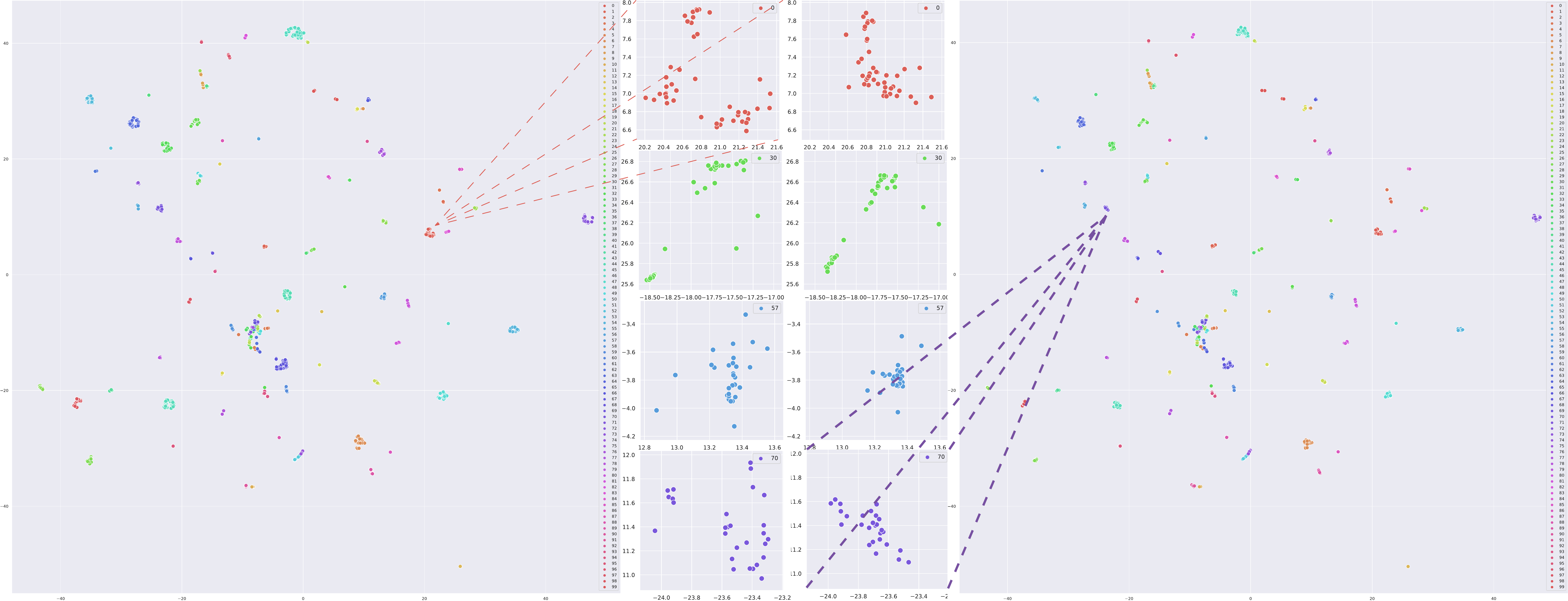}
    \vspace*{-0.58cm}
    \caption{T-SNE visualization of the samples (ID=0,1,...99) on training dataset: Training result of Stage I with $L_{trip}$ (left) v.s. Stage II with $L_{anchor}$ (right). Zoom in to see details.}
    \label{fig:train-tsne}
    \vspace*{-0.68cm}
\end{figure}

\noindent\textbf{Triplet Loss} From the results in \cref{tab:comp-fine-tune}, our methods consistently improve the results over the original model no matter which variant is chosen.
As in \cref{fig:learning-curve}, our methods boost the generalization after the training \textbf{Stage I} converges, where the triplet loss reaches saddle point and there is still room to further intensify the cluster compactness (\cref{fig:train-tsne} left).
After the \textbf{Stage II} training, the intra-class distance is further reduced (\cref{fig:train-tsne} right) and boost the generalization in terms of both rank@1 accuracy and mAP (\cref{tab:comp-fine-tune}).
As illustrated in \cref{fig:view-optimization} and the experiments in \cref{sec:ablation}, our methods perform more effectively after the training \textbf{Stage I} is converged, where a stable feature distribution is provided for aggregation.
On the other hand, triplet loss may provide beneficial effects during the initial stochastic training process.
Triplet loss can inhabit more compacted feature embedding for each class in euclidean space than cross-entropy loss, which has been discussed in BNNeck~\cite{luo-bnneck}.
Consequently, the improvement that the proposed anchor loss brings, is more significant for the model trained with $L_{trip}$ than the one trained without $L_{trip}$ (\cf \cref{tab:comp-fine-tune}\&\cref{fig:learning-curve}).
In summary, anchor loss stimulates stable and effective optimization to find better local optimal when triplet loss suffers from stochastic saddle point (\cf \cref{fig:train-tsne}).

\begin{table}[b]
    \vspace*{-0.68cm}
    \centering
    \resizebox{.45\textwidth}{!}{
        \begin{minipage}{\textwidth}
           \resizebox{\textwidth}{!}{
                \centering
                \begin{tabular}{|c|c c|c c|}
                \hline
                Frequency & \multicolumn{2}{c}{Market1501} & \multicolumn{2}{c}{DukeMTMC reID} \\
                & rank@1 & mAP & rank@1 & mAP \\ \hline
                constant & 95.34\% & 87.91\% & 88.3\% & 78.9\%\\
                epoch    & \textbf{95.37\%} & \textbf{87.99\%} & 88.3\% & \textbf{79.1\%} \\
                iteration & 95.25\%  & 88.11\% & \textbf{88.5\%} & 79.1\% \\ \hline
                \end{tabular}
           }
           \caption{Ablation study for frequency to update anchors}
           \label{tab:comp-frequency}
        \end{minipage}
    }
    \resizebox{.54\textwidth}{!}{
    \begin{minipage}{\textwidth}
        \centering
        \begin{tabular}{|c|c|c|c|c|c|}
            \hline 
            Stage I & Stage II & aggregation method & Rank@1 & mAP \\ \hline
            \multirow{2}{*}{$L_{cls}$} &  - & -  & 93.79\% & 84.69\% \\
            & $L_{cls} + L_{Anchor}$ & avg & 94.18\% & 85.98\%  
            \\ \hline
            \multirow{8}{*}{$L_{cls} + L_{triplet}$}  & - & -  & 94.42\% & 86.18\% \\
            &$L_{cls} + L_{Anchor}$ & avg  & \textbf{95.37\%} & 87.99\% \\
            &$L_{cls} + L_{Anchor}$& weighted  & 95.04\% & 87.95\% \\
            &$L_{cls} + L_{TripletAnchor}$ & avg  & 95.25\% & 87.84\% \\
            &$L_{cls} + L_{TripletAnchor}$ & weighted  & 95.13\% & 87.87\% \\
            &$L_{cls} + L_{Anchor} + L_{triplet}$ & avg  & 95.16\% & 87.87\% \\ 
            &$L_{cls} + L_{TripletAnchor} + L_{triplet}$ & avg & 95.04\% & 87.94\% \\
            &$L_{cls} + L_{TtripletAnchor} + L_{triplet}$ & weighted & 95.28\% & \textbf{88.07\%} \\
            \hline
        \end{tabular}
        \caption{Ablation study of applying anchor loss after the initial training stage, \ie epoch 120.}
        \label{tab:comp-fine-tune}
    \end{minipage}
    }
    \vspace*{-0.38cm}
\end{table}

\noindent\textbf{Applicability} In terms of the comparison about update frequency for anchors, three methods (\cref{alg:anchors-fix},\cref{alg:anchors-epoch},\cref{alg:anchors-iteration}) derive comparable results as shown in \cref{tab:comp-frequency}. Considering the cluster formulation, the anchors generated at the end of traditional training are already well-complied with data distribution regarding their identity labels (\cf \cref{fig:train-tsne} left). After Stage II fine-tuning, the aggregated anchors stay close to the initial one since all the samples are pulled towards their anchors in the optimization (\cf \cref{fig:train-tsne} right). From another perspective, those three methods in \cref{tab:comp-frequency} are alternatives concerning computational cost and training availability while provide comparable result. For example, when training with large training dataset or in the context of online learning, \cref{alg:anchors-fix} and \cref{alg:anchors-iteration} would be preferred with little sacrifice of performance. Hence, our proposed method could be tremendously flexible and widely applicable in terms of training efficiency.  

\begin{table}[t]
\vspace*{-0.18cm}
    \centering
    \scalebox{0.9}{
    \begin{tabular}{c|c c|c c}
         \multirow{2}{*}{Methods}  & \multicolumn{2}{c}{Market1501 $\xrightarrow{}$ DukeMTMC} & \multicolumn{2}{c}{DukeMTMC $\xrightarrow{}$ Market1501} \\
         & Rank@1 & mAP & Rank@1 & mAP \\ \hline
         Resnet50 & 27.9\%(24.3\%) & 15.5\%(13.0\%) & 47.7\%(47.2\%) & 21.7\%(21.1\%) \\
         Resnet50(ours) & 35.3\%(33.3\%) & 20.9\%(18.6\%) & 49.6\%(47.3\%) & 23.0\%(21.8\%) \\
         Resnet50-ibn-a & 40.7\%(37.9\%) & 25.9\%(23.2\%) & 56.0\%(50.6\%) & 27.8\%(24.5\%) \\
         Resnet50-ibn-a(ours) & 46.1\%(46.0\%) & 30.3\%(29.1\%) & 55.3\%(52.0\%) & 28.2\%(25.2\%) \\ 
    \end{tabular}}
    \caption{The performance of different models is evaluated on cross-domain datasets. Market1501 $\xrightarrow{}$ DukeMTMC means that we train the model on Market1501 and evaluate it on DukeMTMC-reID. () denotes the models trained and tested with input size $384\times192$.}
    \label{tab:cross-domain}
    \vspace*{-0.88cm}
\end{table}

\noindent\textbf{Robustness} A natural question about the \textbf{Stage II} fine-tuning is that whether the further cluster-level alignment tends to be domain dependent and overfit the domain distribution.
From the cross-domain testing experiments in \cref{tab:cross-domain}, our methods invariably outperform the baseline model.
It verifies the proposed methods are an effective and robust approach to embed images into identity-related space for metric learning.

\vspace*{-0.18cm}
\subsection{Non-Parametric Anchor vs Parametric Center}
 Although anchors in our work looks like the centers proposed in~\cite{wen-centerloss}, they are intrinsically not the same: the former is non-parametric while the latter is parametric. In fact there is no essential difference between the center loss $\min_{\textbf{c}_j} ||\textbf{f}_{i}-\textbf{c}_j||^2$  and the classification loss $\max_{\textbf{p}_j} (\textbf{f}^{\top}_{i}\textbf{p}_j)$, both of which are distance metrics, \ie $L_2$ distance and inner product. The role played by centers $\textbf{c}_j$ ~\cite{wen-centerloss} corresponds to the role of the hyperplanes $\textbf{p}_j$ in traditional classification. As a result, parametric center loss still conforms to instance-level alignment similar to classification loss under the \textbf{local} mini-batch training framework, and cannot build the \textbf{global} connection in cluster level. On the other hand, in the proposed anchor loss  (\cref{eq:intra-anchor}\&\cref{eq:triplet-anchor}), anchors $\textbf{a}_{j}$ are not optimization variables but calculated from cluster distribution instead. The anchors are iteratively updated from the aggregation of dataset features, which enables them to have the global view of feature distribution during the local mini-batch training.


\begin{figure}[b]
    \vspace*{-0.38cm}
    \centering
    \includegraphics[width=.42\linewidth]{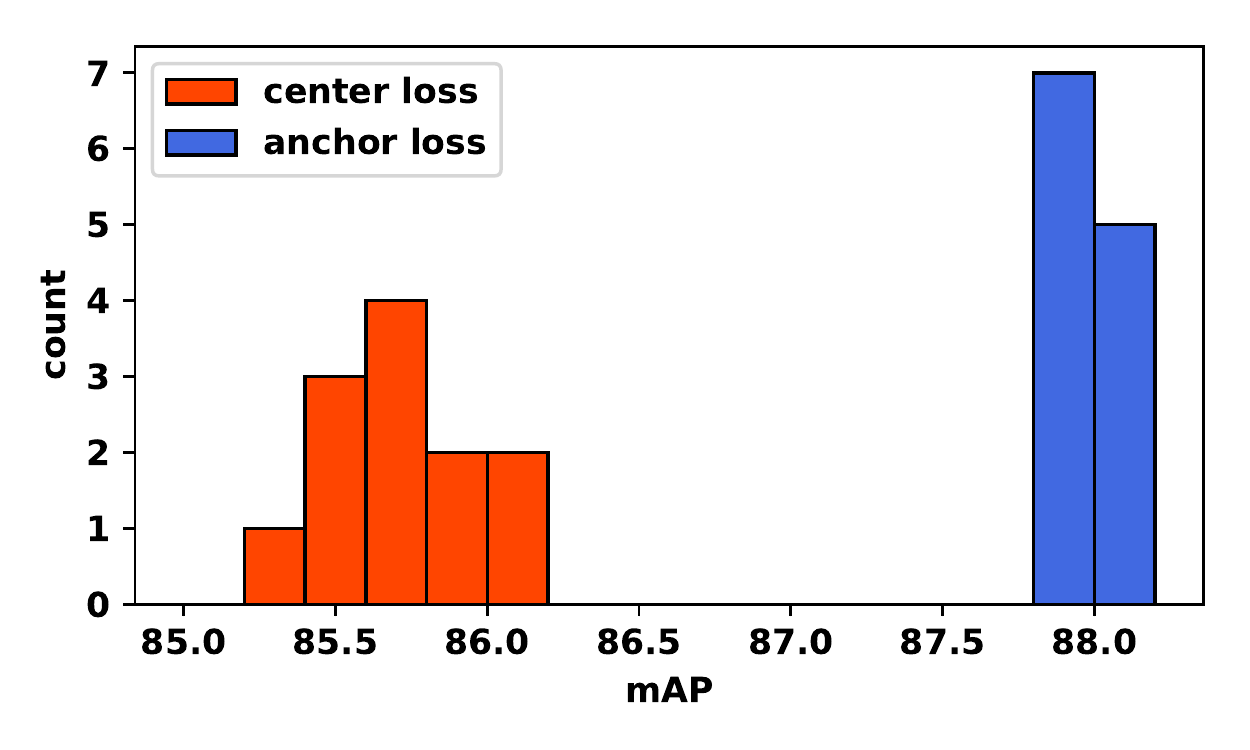}
    \includegraphics[width=.42\linewidth]{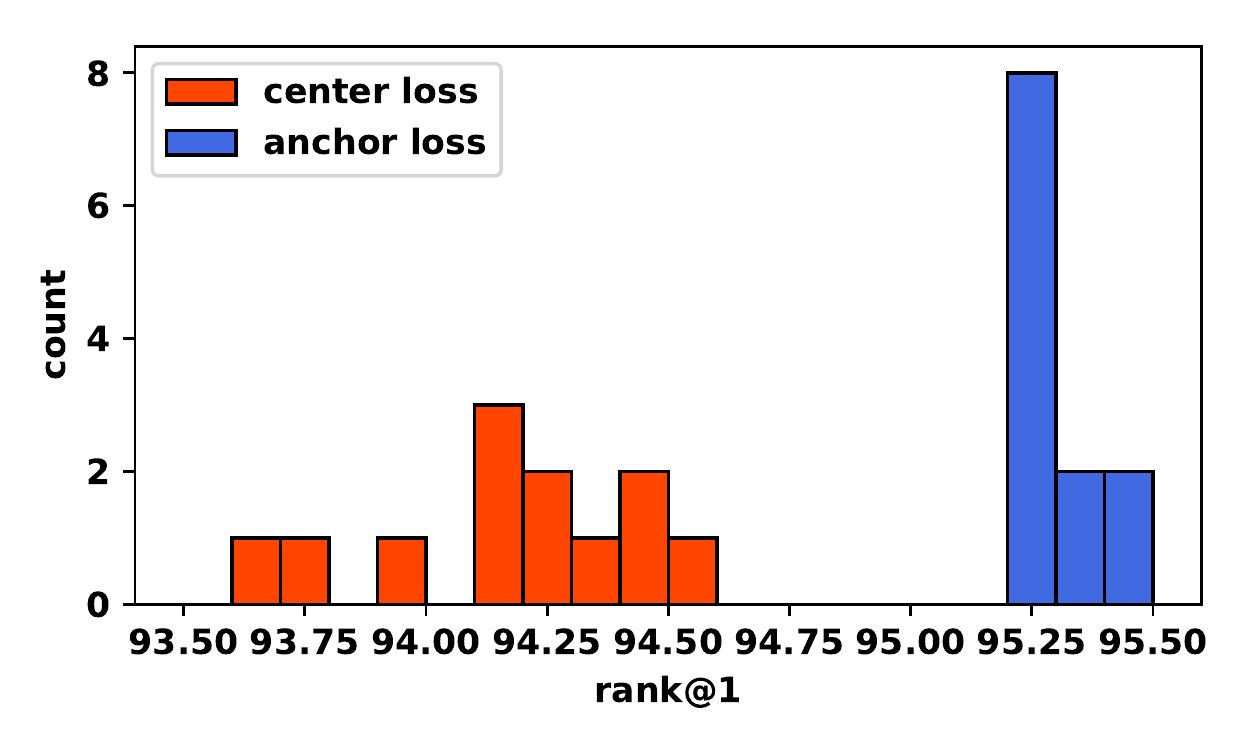}
    \vspace*{-0.58cm}
    \caption{Test result comparison between our method and parametric center loss: We train 12 models independently for each method on Market1501 dataset.}
    \label{fig:comp-initialization}
\end{figure}

We train 12 models independently using the the proposed anchor loss in comparison with another 12 independently trained models using the center loss~\cite{wen-centerloss} on Market1501 dataset.
\cref{fig:comp-initialization} illustrates the model performance histogram in terms of mAP  (left) and rank1 (right). As can be observed, our proposed anchor loss consistently outperform those center loss~\cite{wen-centerloss}. It validates that anchor loss distills the knowledge in latent feature space from the images belonging to the same identity and aggregate them into anchors to guide the training towards well-aligned embedding. Such embedding complies intrinsic feature distribution and thus helps the both training and generalization. Furthermore, the result variance of parametric center loss is much higher than anchor loss, implying the dependency to random initialization for parametric center which may impose some stochastic prior to the cluster formulation. On the contrary, our methods consistently outperform center loss with small variance.

\begin{table}[t]
    \centering
     \scalebox{0.75}{
    \begin{tabular}{c|c|c|c|c}
        \multirow{2}{*}{Methods} & \multicolumn{2}{c}{Labeled} & \multicolumn{2}{c}{Detected} \\ 
        & Rank@1 & mAP & Rank@1 & mAP \\ \hline
        MLFN(CVPR2018)~\cite{chang-multi} & 49.2\% & 54.7\% & 47.8\% & 52.8\% \\
        PCB(ECCV2018)~\cite{sun-pcb} & - & - & 57.5\% & 63.7\% \\
        Mancs(ECCV2018)~\cite{wang-mancs} & 69.0\% & 53.9\% & 65.5\% & 60.5\% \\
        MGN~\cite{wang-mgn} & 67.4\% & 68.0\% & 66.0\% & 68.0\% \\
        Mltb~\cite{yang-cam}(CVPR2019) & 66.5\% & 70.1\% & 64.2\% & 66.6\% \\
        CASN+PCB~\cite{zheng-casn} (CVPR2019) & 73.7\% & 68.0\% & 71.5\% & 64.4\% \\
        MHN-4 (PCB)~\cite{chen-mixedattention}(ICCV2019) & 75.1\% & 70.6\% & 71.6\% & 66.1\% \\ 
        \hline \hline
        Resnet50~\cite{luo-bnneck}* & 63.36\% & 61.60\% & 60.07\% & 51.79\% \\
        Resnet50(ours) & \textbf{76.36\%} & \textbf{74.50\%} & \textbf{72.36\%} & \textbf{70.32\%} \\
        
    \end{tabular}}
    \caption{Comparison with SOTA on CUHK03: CUHK03 evaluation with the setting of 767/700 training/test split on both the labeled and detected images. * denotes our implementation.}
    \label{tab:comp-cuhk}
    \vspace*{-0.38cm}
\end{table}

\begin{table}[t]
\vspace*{-0.18cm}
    \centering
    \resizebox{\linewidth}{!}{
    \begin{tabular}{|l|c|c|c|c|c|}
        \hline 
        \multirow{2}{*}{Methods} & \multirow{2}{*}{alignment method} & \multicolumn{2}{c}{Market-1501} & \multicolumn{2}{c}{DukeMTMC-reID} \\
        & & Rank@1 &  mAP & Rank@1 & mAP \\ \hline
        Mancs (ECCV2018) & - &  93.1\% & 82.3\% & 84.9\% & 71.8\% \\
        PCB+RPP~\cite{sun-pcb} (ECCV2018) & vertical partition & 93.8\% & 81.6\% & 83.3\% & 69.2\% \\
        VPM~\cite{sun-perceive} (CVPR2019) & soft vertical partition & 93.8\% & 80.8\% & 83.6\% & 72.6\% \\ 
        AANet152~\cite{tay-aanet} (CVPR2019) & attribute attention & 93.9\% & 83.4\% & 87.7\% & 74.3\% \\
        IANet~\cite{hou-interaction}(CVPR2019) & spatial semantic & 94.4\% & 83.1\% & 87.1\% & 73.4\% \\
        MltB~\cite{yang-cam} (CVPR2019) & CAM & 94.7\% & 84.5\% & 85.8\% & 72.9 \% \\
        DG-Net~\cite{zheng-joint} (CVPR2019) & latent code & 94.8\% & 86.0\% & 86.6\% & 74.8 \% \\
        MVP Loss~\cite{sun2019mvp} (ICCV2019) & - & 91.4\% & 80.5\% & 83.4\% & 70.0\% \\  
        OSNet~\cite{zhou-omni} (ICCV2019) &  channel attention & 94.8\% & 84.9\% & 88.6\% & 73.5\% \\ 
        MHN-6~\cite{chen-mixedattention} (ICCV2019) & high-order attention & 95.1\% & 85.0\% & 89.1\% & 77.2\% \\
        $P^2$-Net~\cite{guo-part} (ICCV2019) & part & 95.2\% & 85.6\% & 86.5\% & 73.1\% \\
        BDB+Cut~\cite{dai-batch} (ICCV2019) & partition &  95.3\% & 86.7\% & 89.0\% & 76.0\% \\
        ABD-Net~\cite{chen-abd} (ICCV2019) & diverse attention & 95.6\% & 88.3\% & 89.0\% & 78.6\% \\
        \hline \hline 
        Resnet50~\cite{luo-bnneck} & - & 94.1\%(93.6\%) & 85.7\%(85.8\%) &  86.2\%(86.9\%) & 75.9\%(76.8\%) \\
        Resnet50~\cite{luo-bnneck} & parametric center & 94.5\% & 85.9\% & 86.4\% & 76.4\% \\
        Resnet50(ours) & cluster anchor & 95.4\%(94.9\%) & 88.0\% & 88.3\%(89.1\%) & 79.1\%(79.6\%) \\ 
        Resnet50-ibn-a~\cite{luo-bnneck} & - & 95.2\%(95.5\%) & 87.2\%(88.2\%) & 89.0\%(89.7\%) & 79.4\%(80.0\%) \\ 
        Resnet50-ibn-a~\cite{luo-bnneck} & parametric center & 95.0\% & 87.2\% & 89.4\% & 78.8\% \\ 
        Resnet50-ibn-a(ours) & cluster anchor & \textbf{95.7\%(95.8\%)} & \textbf{88.9\%(89.7\%)} & \textbf{90.2\%(91.0\%)} & \textbf{80.6\%(81.8\%)} \\ 
        \hline
    \end{tabular}
    }
    \caption{Comparison of SOTA on Market1501 dataset and DukeMTMC-reID dataset. () denotes the results with a larger input size $384\times192$.}
    \label{tab:comp-sota}
    \vspace*{-0.68cm}
\end{table}

\section{Comparison with the State-of-the-Art Methods}

We compare our method with the recent state-of-the-art methods in \cref{tab:comp-cuhk} and \cref{tab:comp-sota}.
Comparing to spatial alignment by either attention or localization~\cite{sun-pcb,tay-aanet,hou-interaction,yang-cam,guo-part,dai-batch}, our method is much succinct without extra modules or classifier heads to handle subspace alignment.
We have made several attempts to incorporate spatial alignment in our baseline model, \eg PCB~\cite{sun-pcb}, only to find slightly worse results.
Based on our observation from reconstruction experiment (\cf \cref{fig:recon-pipeline}), we notice that decoder can be trained well to reconstruct images in both training and test dataset, implying spatially diversified features are preserved in feature maps before GAP (global average pooling).
However, the feature vectors aggregated after GAP is aligned, meaning GAP could effectively eliminate the spatial variance while preserving the channel-wise diversity when a unified strong model is well trained, \eg the strong baseline model\cite{luo-bnneck}.
Hence, we infer that the benefits of spatial alignment is marginal in our context.
Comparing to channel attention and several variants~\cite{zhou-omni,chen-mixedattention,chen-abd} which aim to generate diverse and uncorrelated feature embedding with the efforts on convolution filters in self-supervised manner, our methods achieve the cluster encoding with focus on feature space by an explicit supervision of aggregated anchors from other images.
DG-Net~\cite{zheng-joint} endeavors to disentangle appearance code and structure code by GAN and auto-encoder, our method accomplishes similar effect that dissects the variance and preserves the identity-related features in the direction towards aggregated anchors (\cf \cref{fig:recon-anchor}).
IBN-Net~\cite{Pan-ibnnet} is proposed to reduce appearance variance and keep discriminative feature extraction by unifying both instance batch normalization and batch normalization.
Luo \etal~\cite{luo-bnneck} apply it as the backbone network for person re-identification and we report our implementation result in \cref{tab:comp-sota}.
We note that resnet50-ibn-a network has the same parameter size and computational cost with original resnet50.
Without bells and whistles, our method consistently boosts the performance in terms of both Rank@1 and mAP comparing to corresponding baseline (resnet50 and resnet50-ibn-a~\cite{luo-bnneck}), achieving the state-of-the-art results on Market-1501, DukeMTMC-reID(\cref{tab:comp-sota}) and CUHK03(\cref{tab:comp-cuhk}) datasets..
Specially, due to further reduce of intra-class variance towards a compact cluster in latent feature space, our method improves mAP significantly and benefits the robustness for the application of person re-identification.
In summary, comparing to the recent state-of-the-art methods, our methods visit another modality of alignment, cluster-level alignment, validating that exploration of interaction of clusters observed from dataset feature distribution improves both training and generalization.

\section{Conclusion}

In this paper, we investigate the person re-identification from the view of alignment and find an interesting and effective approach to delve in another scale of feature alignment, cluster level.
By performing aggregation and alignment iteratively, our proposed anchor loss is enabled to interact with more images indirectly from aggregated anchors, which pass the distilled knowledge from the feature distribution and provide a consistent optimization to further boost the performance significantly after traditional training convergence.
It shows the cluster-level alignment guided by the aggregation of dataset distribution, which steps beyond the general classification framework, is essential and beneficial for identity-related embedding.

\clearpage
%
%
\bibliographystyle{splncs04}
\bibliography{reid}
\end{document}